\pdfoutput=1
\documentclass[12pt]{article}
\usepackage[T1]{fontenc}
\usepackage{graphicx}
\usepackage{sbc-template}
\usepackage{subfig}
\usepackage{booktabs}
\usepackage{xcolor}
\usepackage{adjustbox}
\usepackage{hyperref}
\usepackage{multirow}
\usepackage{amsmath}
\usepackage{amssymb}
\usepackage[flushleft]{threeparttable}

\title{
Closed-Loop Evaluation of Bird’s-Eye-View Maps from Cross-View Transformers as Inputs to Behavior-Cloning Policies
}

\author{
Felipe Carlos dos Santos\inst{1}, 
Eric Aislan Antonelo\inst{1}, 
Gustavo Claudio Karl Couto\inst{1}
}
\address{
Federal University of Santa Catarina (UFSC), Florianópolis, Brazil \\
Department of Automation and Systems Engineering 
\email{felipe.carlos.santos@posgrad.ufsc.br}, 
\email{eric.antonelo@ufsc.br, gustavo.karl.couto@posgrad.ufsc.br}
}
\begin{document}
\maketitle
\begin{abstract}
In autonomous driving, Bird's-Eye View (BEV) representations provide a structured, top-down abstraction of the vehicle's surroundings and have become a key input modality for Behavioral Cloning (BC) policies. While ground-truth BEV maps are readily available in simulation, real-world deployment requires replacing them with camera-predicted counterparts—a substitution that introduces perceptual errors whose downstream impact on closed-loop driving performance is not well understood. In this work, we investigate the use of Cross-View Transformer (CVT)-predicted BEV maps as direct policy inputs for a BC agent in the CARLA simulator. We propose a six-channel BEV representation covering road surface, planned route, lane boundaries, vehicles, pedestrians, and traffic lights, and introduce a Kernel Density Estimation (KDE) weighting scheme that rebalances the segmentation loss towards underrepresented driving maneuvers such as curves and intersections. Closed-loop evaluation across two CARLA towns shows that the KDE-weighted model is the only predicted-BEV agent to complete a full episode without infractions, despite not achieving the highest aggregate IoU. This discrepancy reveals that global segmentation metrics are poor proxies for driving performance: what determines navigation success is prediction quality at geometrically critical locations, and the route channel emerges as the primary bottleneck for reliable agent navigation under predicted BEV inputs.

\end{abstract}

\section{Introduction}

Autonomous driving has traditionally relied on modular pipelines that sequentially process perception, planning, and control. However, learning-based approaches have gained significant traction for their ability to capture complex driving behaviors directly from data, often leveraging Behavior Cloning (BC) trained on offline expert demonstrations \cite{nvidiabc,chauffernet,codevilla2018endtoend}. A critical enabler for these methods is the Bird's-Eye View (BEV) representation, which removes perspective distortions from ego-centric cameras and exposes scene geometry in a planner-friendly grid~\cite{philion2020lss}. Recent transformer-based architectures learn multi-camera view transformations implicitly: Cross-View Transformers (CVT)~\cite{park2023cvt} introduced camera-aware cross-view attention for real-time semantic BEV generation from onboard cameras alone, without requiring LiDAR or dense depth supervision, making BEV prediction feasible in camera-only, single-vehicle settings.

Decoupling perception from policy learning through structured BEV representations simplifies the control task and improves sample efficiency \cite{resnet_rl,Couto2023}. Behavior cloning (BC) from limited expert demonstrations can be effectively conditioned on BEV representations and sparse route inputs for simulated city navigation~\cite{antonelo2024investigating}, validating BEV as a compact, semantically rich interface between perception and control. Preliminary studies have explored CVT-based BEV generation under single-vehicle configurations and limited training towns~\cite{santos2025initial}, yet the downstream impact of camera-only predicted BEVs on closed-loop driving performance remains not adequately investigated.

This work proposes BEV generation models trained to predict six one-hot encoded semantic channels: roads, planned route, lane boundaries, vehicles, pedestrians, and traffic lights. To assess the practical utility of these representations, the generated BEVs are integrated into a BC driving policy and their downstream impact on autonomous navigation is evaluated. As an initial benchmark, the proposed models are compared against the three-channel baselines~\cite{santos2025initial} in a simulated driving environment without dynamic obstacles, enabling an isolated analysis of the agent's ability to follow routes and execute driving maneuvers using predicted BEV representations.
The main contributions of this work are:
\begin{itemize}
    \item A six-channel CVT-based BEV generation model trained on a large, perturbation-augmented dataset, with six channels in contrast to the three-channel model of~\cite{santos2025initial}, encoding roads, route, lane boundaries, vehicles, pedestrians, and traffic lights.
    \item A KDE-based importance-weighting scheme that rebalances the BEV segmentation loss towards underrepresented driving maneuvers, such as curves and intersections, improving route and road channel quality at geometrically critical locations.
    \item A closed-loop evaluation of multiple BEV generation models as policy inputs in the CARLA simulator.
\end{itemize}

\section{Methods}
\label{sec:methods}


%
%
%
%
%
%

\subsection{Cross-view transformer (CVT)}
\label{sec:cvt_exp}

CVT~\cite{park2023cvt} processes a set of $n$ monocular views $\{(I_k, K_k, R_k, t_k)\}_{k=1}^n$, where $I_k \in \mathbb{R}^{H \times W \times 3}$ is the image, $K_k \in \mathbb{R}^{3 \times 3}$ the intrinsic matrix, and $(R_k, t_k)$ the extrinsic rotation and translation relative to the ego-vehicle. Images are encoded by an EfficientNet-B4~\cite{tan2019efficientnet} backbone, and camera-aware positional encodings derived from per-view projection geometry are fused with image features via cross-view attention, aggregating multi-view information into a unified BEV latent space. The map-view representation is iteratively refined at multiple resolutions before being upsampled to the final prediction. Full details on the attention formulation and positional encoding can be found in~\cite{park2023cvt}.

The CVT output is a six-channel $192{\times}192$ top-down map, where each channel is a binary segmentation mask for one semantic class: road surface, planned route, lane boundaries, vehicles, pedestrians, and traffic lights.


\subsection{Behavior Cloning (BC)}
\label{sec:bc}

Behavior Cloning (BC) trains a policy $\pi_\theta$ on a dataset $\mathcal{D} = \{(s_i, a_i)\}_{i=1}^N$ of expert demonstrations collected in CARLA, where $s_i$ pairs a ground-truth BEV image with the vehicle's current speed and previous control commands, and $a_i = (a_i^{\mathrm{acc}}, a_i^{\mathrm{steer}}) \in [-1,1]^2$ is the expert's acceleration and steering~\cite{pomerleau1988alvinn}. Given $s_i$, the network outputs four scalars $(\alpha^{\mathrm{acc}}, \beta^{\mathrm{acc}}, \alpha^{\mathrm{steer}}, \beta^{\mathrm{steer}})$ that parameterize independent Beta distributions; at inference, control actions are sampled from these distributions~\cite{Petrazzini2021,antonelo2024investigating}. The bounded support of $\mathcal{B}(\alpha,\beta)$ on $[0,1]$ naturally accommodates actuator constraints without post-hoc clipping. Actions on $[-1,1]$ are rescaled via $a' = (a+1)/2$, and the training objective is the mean negative log-likelihood:
\begin{equation}
  \mathcal{L}(\theta)
  = -\frac{1}{N}\sum_{i=1}^{N}
    \left[
      \log f\!\left(\tfrac{a_i^{\mathrm{acc}}+1}{2};\;
                    \alpha_i^{\mathrm{acc}},\beta_i^{\mathrm{acc}}\right)
      +
      \log f\!\left(\tfrac{a_i^{\mathrm{steer}}+1}{2};\;
                    \alpha_i^{\mathrm{steer}},\beta_i^{\mathrm{steer}}\right)
    \right],
  \label{eq:bcloss_car}
\end{equation}
where $f(x;\alpha,\beta) = x^{\alpha-1}(1-x)^{\beta-1}/B(\alpha,\beta)$ is the Beta density and $B(\alpha,\beta)$ is the Beta function.

\subsection{Kernel Density Estimation for Segmentation Training}
\label{sec:kde}

A Kernel Density Estimator (KDE) is a non-parametric method to estimate the probability density function (PDF) of a random variable.
For each data point in the sample, a kernel function is centered at that point. The KDE is computed by summing all these kernel functions and normalizing by the number of data points.
Mathematically, the KDE for a point $x$ is given by:

\begin{equation}
   \hat{g}(x) = \frac{1}{N h} \sum_{j=1}^{N} K \left( \frac{x - x_j}{h} \right),
    \label{eq:kde}
\end{equation}
where $\hat{g}(x)$ is the estimated density at point $x$, $N$ is the number of data points, $h$ is the bandwidth, $x_j$ are the data points, and $K$ is the kernel function.

The Gaussian kernel is commonly used in KDE as it assigns weights to data points based on their distance from the estimation point, ensuring that closer points exert a greater influence on the density estimate. It is defined as:

\[ K(u) = \frac{1}{\sqrt{2\pi}} e^{-\frac{u^2}{2}}, \]
where $u$ is the standardized distance between the estimation point and the data point, scaled by the bandwidth $h$.

In this work, KDE is applied to reweight training samples during the optimization of one of our studied  BEV segmentation models.
The KDE is fitted over the trajectory actions (steering angle and throttle) recorded in the training set, providing a density estimate of the action distribution.
By using the inverse of this density as a per-sample weight, the model is encouraged to pay greater attention to underrepresented driving scenarios, such as sharp turns and intersections, which are the most challenging and least frequent contexts in the training data.

Specifically, the per-sample weight $w_i$ is defined as:
\begin{equation}
    w_i = \frac{1 / \hat{g}(a_i)}{\frac{1}{N}\sum_{j=1}^{N} 1/\hat{g}(a_j)},
    \label{eq:kde_weight}
\end{equation}
where $\hat{g}(a_i)$ is the kernel density estimate evaluated at the action $a_i$ of sample $i$, and the denominator normalizes the weights to have unit mean across the training set.

The per-sample segmentation loss $\ell_i$ is the binary cross-entropy (BCE) averaged over all $C$ output channels and the spatial dimensions $H \times W$ of the BEV map:
\begin{equation}
    \ell_i = \frac{1}{C \cdot H \cdot W} \sum_{c=1}^{C} \sum_{h,w}
    \mathrm{BCE}\!\left(\hat{y}_{i,c,h,w},\; y_{i,c,h,w}\right),
\end{equation}
where $\hat{y}_{i,c,h,w}$ is the model prediction and $y_{i,c,h,w}$ is the ground-truth binary mask for channel $c$ at spatial position $(h, w)$.

The resulting KDE-weighted training objective is:
\begin{equation}
    \mathcal{L}(\theta) = \frac{1}{N} \sum_{i=1}^{N} w_i \cdot \ell_i.
    \label{eq:weighted_loss}
\end{equation}

This formulation ensures that samples from rare driving situations contribute more strongly to the gradient, thereby biasing the model towards better performance on the scenarios where accurate BEV prediction matters most.

\section{BC Policy Agent}
\label{sec:agent}
The BC policy agent takes a BEV image and scalar vehicle state as input, and outputs continuous control actions via a Beta distribution head. The following subsections detail the input representation, action space, and network architecture.

\subsection{Input representation}
The policy agent receives as input a $192 \times 192$ three-channel Bird's-Eye View (BEV) image produced by the perception network. In the present evaluation, only the first three semantic channels are employed, corresponding to the road surface, planned route, and lane-boundary representations (channels 0--2). This restriction is motivated by three factors: it enables direct comparison with the three-channel baselines from~\cite{santos2025initial}; the KDE weighting strategy described in Section~\ref{sec:kde} improves prediction quality primarily for these static geometry channels, since they are most affected by the underrepresented driving maneuvers the KDE targets; and channels 3--5 (vehicles, pedestrians, and traffic lights) are severely class-imbalanced relative to channels 0--2: whereas road, route, and lane-boundary pixels cover large and consistent fractions of each frame, dynamic elements occupy sparse, intermittent regions, posing a representational challenge that translates directly into near-zero IoU across all trained models, as further discussed in Section~\ref{sec:experiments} (Table~\ref{tab:iou_test_town02}); their improvement is left for future work.

In addition to the visual input, the agent receives the current longitudinal speed of the ego-vehicle along with the most recent policy actuator outputs, namely the previous acceleration and steering commands. These scalar inputs are concatenated with the feature representation at a later stage of the network, specifically at the first fully connected layer.

\subsection{Output Representation}

The action space is $\mathbf{a} \in [-1, 1]^{2}$, comprising steering and a unified acceleration signal that encodes braking as negative acceleration, preventing simultaneous throttle and brake commands~\cite{Petrazzini2021}. The policy output is modeled by a Beta distribution as described in Section~\ref{sec:bc}.

\subsection{Network architecture}

The BC agent employs the same network architecture used in~\cite{antonelo2024investigating}.
The BEV image is processed through a stack of convolutional layers, producing a spatial embedding that is concatenated with a state embedding obtained by passing the scalar vehicle state variables through a small MLP.
The fused representation is then projected to a 256-dimensional feature vector, which is mapped by a subsequent MLP policy head to the parameters of a Beta distribution, from which the continuous control actions, acceleration \(a^{\mathrm{acc}} \in [0, 1]\) and steering \(a^{\mathrm{steer}} \in [-1, 1]\), are sampled.

\section{Experiments}
\label{sec:experiments}

In this section, we evaluate three CVT models—\textit{CVT 6ch.\ No Noise}, \textit{CVT 6ch.}, and \textit{CVT 6ch.\ KDE}—first via per-channel IoU for BEV quality, then via closed-loop driving in CARLA to assess the downstream impact of predicted BEV on autonomous navigation.

\subsection{Dataset Generation}
\label{sec:dataset}
The simulation
environment and route definitions are sourced from the CARLA Leaderboard evaluation
platform~\cite{leaderboard_2020}. Specifically, the \textit{Town01} environment is
employed for dataset collection, comprising ten predefined routes used to build the
expert demonstration dataset, while \textit{Town02} is reserved exclusively for
testing purposes. Each town provides ten distinct routes.

Compared to the dataset from~\cite{santos2025initial}, a larger and more diverse dataset is proposed
here, as the inclusion of dynamic agents, vehicles, pedestrians, and traffic
lights, substantially increases the complexity of the driving task.
The dataset consists of five independent collections of the ten routes available
in \textit{Town01}. Across all five collections, route 00 was designated as the
validation split during training, while routes 01 through 09 constituted the
training set.

To increase data variability and simulate different levels of driving proficiency,
steering and throttle perturbations were injected at varying magnitudes, ranging
from large disturbances capable of displacing the vehicle into the opposing lane or
off the road entirely, to zero-noise demonstrations. Additionally, each collection
was generated using a distinct random seed, resulting in different combinations of
vehicle and pedestrian appearances, shapes, and spawn configurations across
collections.  

In total, the dataset comprises 1{,}186{,}499 images and 201{,}980 sample points (${\approx}$93.8~GB):

\begin{itemize}
    \item \textbf{Training:} 930{,}040 images, 155{,}007 points (78.2~GB).
    \item \textbf{Validation:} 129{,}579 images, 21{,}597 points (10.5~GB).
    \item \textbf{Test:} 126{,}880 images, 25{,}376 points (5.1~GB).
\end{itemize}

Compared to the dataset from~\cite{santos2025initial}, the training and validation sets grew approximately 10$\times$ (from 106{,}435 to 1{,}059{,}619 images; from 4.1~GB to 88.7~GB), while the test set remains unchanged to preserve comparability with previously reported results.


\subsection{Training}

Three models were trained using the Binary Cross-Entropy (BCE) loss with six input
channels: four RGB images captured from front, right, left, and rear-facing cameras,
and one single-channel image representing a sparse route map, following the
architecture described in Section~\ref{sec:cvt_exp}. All networks presented in this paper were trained for 50 epochs, following the same setting used in~\cite{santos2025initial}, to ensure a uniform comparison across models.

\textit{CVT 6ch.\ No Noise} uses the smaller, unperturbed dataset from~\cite{santos2025initial}; the 3-channel vs.\ 6-channel distinction is relevant when comparing against that baseline. \textit{CVT 6ch.} uses the same architecture on the larger dataset from Section~\ref{sec:dataset}. \textit{CVT 6ch.\ KDE} adds the importance-weighting scheme from Section~\ref{sec:kde}, assigning higher BCE loss weights to underrepresented curve and intersection samples.

The BC policy is trained exclusively with ground-truth BEV maps obtained directly
from CARLA. This choice avoids retraining a separate policy for each BEV generation
model, which would introduce an additional source of variability and make it harder
to isolate the effect of BEV prediction quality on driving performance. At evaluation
time, the ground-truth BEV is replaced by the output of each perception model, so
that the only varying factor across runs is the BEV generator.

\subsection{Evaluation Protocol}

The evaluation is divided into two parts. First, BEV prediction quality is assessed using Intersection over Union (IoU) computed independently for each output channel, on the validation split of Town01 and on the held-out test set of Town02. Second, the driving performance of the BC agent is measured in closed-loop episodes within the CARLA simulator, using the generated BEV maps as policy input in place of the ground-truth maps used during training.

\subsection{BEV Evaluation}

Table~\ref{tab:iou_val_town01} shows that the six-channel BEV task splits into two sub-problems: channels 0--2 (roads, route, lane boundaries) encode static geometry present in every frame, while channels 3--5 (vehicles, pedestrians, traffic lights) require dynamic obstacle recognition, a harder task given element sparsity in the data.

For channels 0--2, KDE weighting provides a clear benefit: \textit{CVT 6ch.\ KDE} achieves the highest mean IoU (0.6413) among the larger-dataset models, outperforming \textit{CVT 6ch.} (0.5897). Since curve and intersection samples receive higher loss weight, and these geometries directly determine road, route, and lane-boundary quality, the improvement is consistent with the KDE motivation. \textit{CVT 6ch.\ No Noise}, trained on the smaller unperturbed dataset from~\cite{santos2025initial}, achieves only 0.2455, confirming that dataset scale and perturbation diversity are key drivers of segmentation quality independently of KDE weighting.

\begin{table}[ht]
\centering
\caption{Mean per-channel IoU — City 1 (Town01, validation set).}
\label{tab:iou_val_town01}
\begin{threeparttable}
\begin{tabular}{lccc}
\toprule
\textbf{Channel}
  & \textbf{CVT 6ch.}
  & \textbf{CVT 6ch. No Noise}
  & \textbf{CVT 6ch. KDE} \\
\midrule
Roads          & 0.8135 & 0.5086 & \textbf{0.8614} \\
Trajectory     & 0.5938 & 0.1918 & \textbf{0.6455} \\
Lane           & 0.3618 & 0.0362 & \textbf{0.4170} \\
Vehicles       & \textbf{0.2460} & 0.0696 & 0.2401 \\
Pedestrians    & 0.0000 & 0.0000 & 0.0000 \\
Traffic Light  & \textbf{0.0532} & 0.0031 & 0.0499 \\
\midrule
Mean (Ch.\ 0--2) & 0.5897 & 0.2455 & \textbf{0.6413} \\
Mean (Ch.\ 3--5) & \textbf{0.0997} & 0.0242 & 0.0967 \\
\bottomrule
\end{tabular}
\begin{tablenotes}
\small
\item Bold values indicate the best score per row.
\end{tablenotes}
\end{threeparttable}
\end{table}


For the obstacle-related channels (3--5), however, the benefit of KDE weighting is not observed. The mean IoU for channels 3--5 is comparable between \textit{CVT 6ch.} (0.0997) and \textit{CVT 6ch.\ KDE} (0.0967), with the KDE variant even showing a marginal decrease relative to the vanilla model. This suggests that reweighting samples based on the density of steering and throttle actions, which is informative for the static road geometry task, does not transfer to the detection of vehicles, pedestrians, and traffic lights, whose occurrence is governed by factors largely independent of the ego-vehicle's own control actions. Addressing this limitation is left for future work.

The results on the test set, summarized in Table~\ref{tab:iou_test_town02}, corroborate these findings in an unseen town. The \textit{CVT 6ch.\ KDE} model again achieves the best mean IoU for channels 0--2 (0.5667) among the models trained on the proposed dataset, surpassing both \textit{CVT 6ch.} (0.5519) and \textit{CVT 6ch.\ No Noise} (0.4459), while remaining below the \textit{CVT 3ch.} baseline (0.6703). For channels 3--5, the pattern observed in Town01 persists, with \textit{CVT 6ch.} and \textit{CVT 6ch.\ KDE} performing similarly (0.0887 and 0.0852, respectively), reinforcing that the advantage of KDE weighting is confined to the road-geometry sub-task and does not generalize to obstacle recognition. 

It is worth noting, however, that route-averaged IoU is inherently biased towards straight segments, which dominate route length, while curves and intersections—driving-critical locations—account for a small fraction of the total. This explains why aggregate metrics favor \textit{CVT 3ch.} (Table~\ref{tab:iou_spawns}), yet this advantage does not translate into better closed-loop driving performance, as discussed in the next section.

\begin{table}[ht]
\centering
\caption{Mean per-channel IoU — City 2 (Town02, test set).}
\label{tab:iou_test_town02}
\begin{threeparttable}
\begin{tabular}{lccc|cc}
\toprule
\textbf{Channel}
  & \textbf{CVT 6ch.}
  & \textbf{CVT 6ch. No Noise}
  & \textbf{CVT 6ch. KDE}
  & \textbf{CVT 3ch.}$^{\dagger}$
  & \textbf{UNet}$^{\dagger}$ \\
\midrule
Roads          & 0.7889 & 0.6810 & 0.8160 & \textbf{0.9144} & 0.6978 \\
Trajectory     & 0.5569 & 0.4082 & 0.5755 & \textbf{0.7808} & 0.5915 \\
Lane           & 0.3098 & 0.2485 & 0.3086 & 0.3158          & \textbf{0.3959} \\
Vehicles       & \textbf{0.2419} & 0.1221 & 0.2252 & --     & --     \\
Pedestrians    & 0.0001 & 0.0000 & \textbf{0.0003} & --     & --     \\
Traffic Light  & 0.0240 & 0.0051 & \textbf{0.0302} & --     & --     \\
\midrule
Mean (Ch.\ 0--2) & 0.5519 & 0.4459 & 0.5667 & \textbf{0.6703} & 0.5617 \\
Mean (Ch.\ 3--5) & \textbf{0.0887} & 0.0424 & 0.0852 & --     & --     \\
\bottomrule
\end{tabular}
\begin{tablenotes}
\small
\item Bold values indicate the best score per row.
\item[$\dagger$] Models trained with only 3 output channels in \cite{santos2025initial}.
\end{tablenotes}
\end{threeparttable}
\end{table}

\subsection{BC Agent Evaluation}

All BEV generation models are evaluated by substituting the ground-truth BEV with their respective predictions at inference time, while keeping the policy weights fixed. To isolate the effect of BEV prediction quality on driving performance, the evaluation environment contains no dynamic agents: other vehicles and pedestrians are absent, and traffic lights are disabled. Under these conditions, the policy driven by the ground-truth BEV successfully completes all five episodes in both towns without any infraction (Table~\ref{tab:spawn_route_completion}), establishing a strong upper-bound reference for the evaluation.

\begin{figure}[ht]
    \centering

    
        \includegraphics[width=1\linewidth]{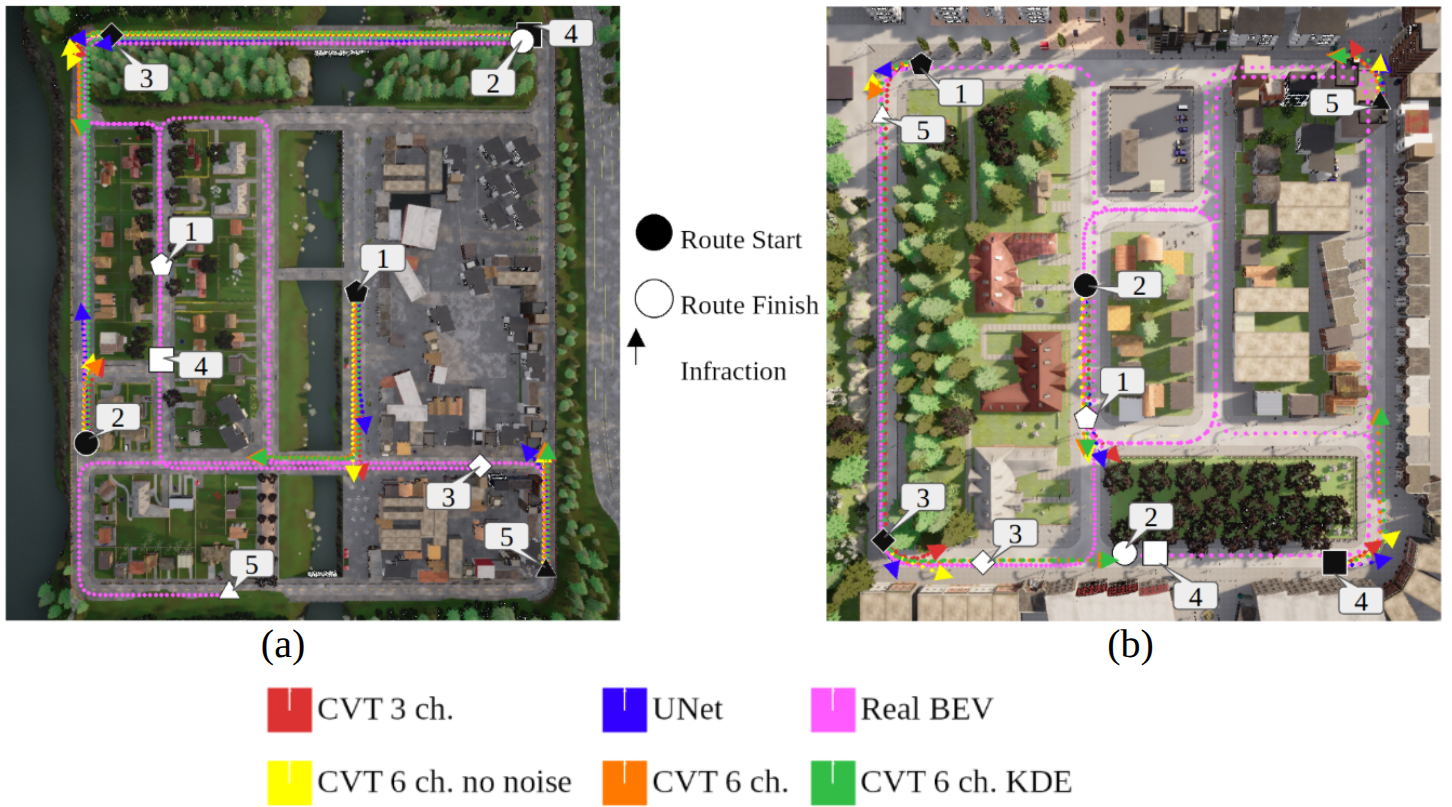}

    \caption{Trajectories generated by each evaluated model overlaid on the CARLA top-down maps for (a) Town01 and (b) Town02. Colors denote different BEV generation approaches, as indicated in the legend. Black filled markers represent episode starting locations, while white filled markers indicate successfully completed episodes. Colored arrows mark the vehicle position and heading at the time of an infraction that caused episode termination. The numerical labels identify the corresponding episode start and end points.}
    \label{fig:trajectories}
\end{figure}

Each model was evaluated in five fixed-spawn closed-loop episodes per town, each defining a route of approximately 1{,}241~m; episodes end on route completion or upon an infraction. Figure~\ref{fig:trajectories} overlays the resulting trajectories on the CARLA top-down maps: colored paths trace each model's route, arrows mark infraction positions, and numbered markers link episode start and end points.

\begin{figure}[ht]
    \centering
   \begin{minipage}{0.24\textwidth}
    \centering
    \begin{tabular}{@{}c@{}}
        \includegraphics[width=\linewidth]{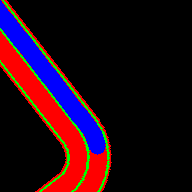} \\
        \textbf{Ground Truth}
    \end{tabular}
\end{minipage}
    \hfill
    \begin{minipage}{0.75\textwidth}
        \centering
        \begin{tabular}{@{}ccc@{}}
            \includegraphics[width=0.3\linewidth]{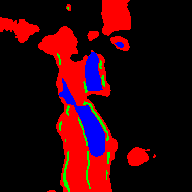} &
            \includegraphics[width=0.3\linewidth]{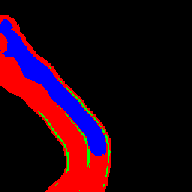} &
            \includegraphics[width=0.3\linewidth]{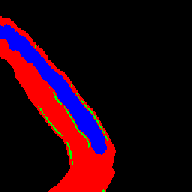} \\
             CVT 3ch. &  CVT 6ch. KDE &  CVT 6ch.
        \end{tabular}
    \end{minipage}

    \caption{BEV predictions at the northwest curve of \textit{Town01} near the start of episode three, where the road transitions from an east--west straight to a southward turn.
predictions at the same timestep. Only the CVT~6ch. and CVT~6ch.~KDE agents complete this curve.
}

    \label{fig:curva_3_town01}
\end{figure}

Table~\ref{tab:spawn_route_completion} summarizes the route completion distances and infraction counts for each model across both towns. Among all models that rely on predicted BEV representations, \textit{CVT 6ch.\ KDE} achieves the best overall driving performance, accumulating 261~m of route completion in Town01 and 106~m in Town02. More notably, it is the only perception model capable of completing an entire episode without triggering any infraction, recording four infractions out of five episodes in Town01 rather than the five out of five observed for every other predicted-BEV model. 

As shown in Figure~\ref{fig:trajectories}, the primary failure modes of the evaluated agents occur at curves and road intersections. These are precisely the scenarios where KDE-based weighting proves beneficial. By assigning greater importance to training samples associated with turning maneuvers and intersection navigation, \textit{CVT 6ch. KDE} learns more robust representations for these underrepresented situations, resulting in more reliable BEV predictions and improved route completion performance.


\begin{table}[htbp]
\centering
\caption{Route completion (meters) and number of infractions
         out of 5 episodes for Town01 and Town02
         (route length ${\approx}$~1{,}241~m per town).}
\label{tab:spawn_route_completion}
\begin{tabular}{l rr rr}
\toprule
 & \multicolumn{2}{c}{\textbf{Town01}} & \multicolumn{2}{c}{\textbf{Town02}} \\
\cmidrule(lr){2-3}\cmidrule(lr){4-5}
\textbf{BEV Generator} & Route (m) & Infrac. & Route (m) & Infrac. \\
\midrule
\textit{Ground-truth BEV} & \textit{598} & \textit{0/5} & \textit{604} & \textit{0/5} \\
\midrule
UNet             &  115 & 5/5 &   22 & 5/5 \\
CVT 3ch.         &  159 & 5/5 &   27 & 5/5 \\
CVT 6ch.         &  151 & 5/5 &   52 & 5/5 \\
\textbf{CVT 6ch. KDE} & \textbf{261} & \textbf{4/5} & \textbf{106} & \textbf{5/5} \\
CVT 6ch. No Noise &  120 & 5/5 &   28 & 5/5 \\
\bottomrule
\end{tabular}
\end{table}

\begin{figure}[ht]
    \centering
   \begin{minipage}{0.24\textwidth}
    \centering
    \begin{tabular}{@{}c@{}}
        \includegraphics[width=\linewidth]{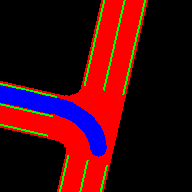} \\
        \textbf{Ground Truth}
    \end{tabular}
\end{minipage}
    \hfill
    \begin{minipage}{0.75\textwidth}
        \centering
        \begin{tabular}{@{}ccc@{}}
            \includegraphics[width=0.3\linewidth]{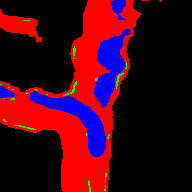} &
            \includegraphics[width=0.3\linewidth]{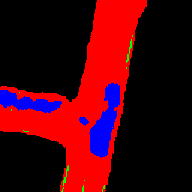} &
            \includegraphics[width=0.3\linewidth]{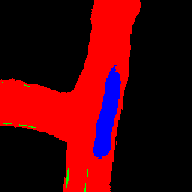} \\
              CVT 3ch. &  CVT 6ch. KDE &  CVT 6ch.
        \end{tabular}
    \end{minipage}

    \caption{BEV predictions at the cross road in the middle of \textit{Town02} near the finish point of episode one.
predictions at the same timestep. All agents fail in this intersection.
}
    \label{fig:cruzamento_1_town02}
\end{figure}

The episode successfully completed by \textit{CVT 6ch.\ KDE} in Town01 corresponds to the second episode in Figure~\ref{fig:trajectories}(a). This route requires the agent to navigate through at least one curve and two intersections before reaching the finish point. The fact that \textit{CVT 6ch.\ KDE} completes this episode while all other models fail underscores a key finding: a higher IoU mean on the evaluation routes does not necessarily translate into better closed-loop driving performance. As shown in Table~\ref{tab:iou_spawns}, \textit{CVT 3ch.} achieves the highest mean IoU in both towns (0.764 in Town01 and 0.679 in Town02), yet it fails all five episodes in each town with the shortest average route completion. Conversely, \textit{CVT 6ch.\ KDE} ranks second in mean IoU (0.713 in Town01) but delivers the strongest driving results. This discrepancy suggests that the KDE weighting scheme, by up-weighting training samples associated with curves and intersections, produces a BEV representation that is more reliable precisely at the critical decision points where the policy is most likely to commit an infraction, even if this does not manifest as a global improvement in aggregate IoU metrics.

\begin{table}[ht]
\centering
\caption{Mean IoU per channel on the spawn evaluation for Town01 and Town02.
Channels: Road (ch.\,0), Route (ch.\,1), Lane boundary (ch.\,2).}
\label{tab:iou_spawns}
\resizebox{\linewidth}{!}{%
\begin{tabular}{lcccc cccc}
\toprule
& \multicolumn{4}{c}{\textbf{Town01}} & \multicolumn{4}{c}{\textbf{Town02}} \\
\cmidrule(lr){2-5}\cmidrule(lr){6-9}
\textbf{BEV Generator} & Road & Traj. & Lane & \textbf{Mean} & Road & Traj. & Lane & \textbf{Mean} \\
\midrule
CVT 3ch.              & 0.933 & \textbf{0.808} & \textbf{0.551} & \textbf{0.764} & 0.871 & \textbf{0.727} & \textbf{0.438} & \textbf{0.679} \\
CVT 6ch.               & 0.935 & 0.788 & 0.332 & 0.685 & \textbf{0.903} & 0.680 & 0.321 & 0.635 \\
CVT 6ch. KDE           & \textbf{0.943} & 0.789 & 0.406 & 0.713 & 0.902 & 0.667 & 0.340 & 0.636 \\
\bottomrule
\end{tabular}%
}
\end{table}

\begin{figure}[ht]
    \centering
   \begin{minipage}{0.24\textwidth}
    \centering
    \begin{tabular}{@{}c@{}}
        \includegraphics[width=\linewidth]{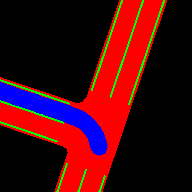} \\
        \textbf{Ground Truth}
    \end{tabular}
\end{minipage}
    \hfill
    \begin{minipage}{0.75\textwidth}
        \centering
        \begin{tabular}{@{}cc@{}}

            \includegraphics[width=0.3\linewidth]{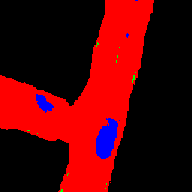} &
            \includegraphics[width=0.3\linewidth]{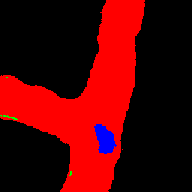} \\
              CVT 6ch. KDE &  CVT 6ch.
        \end{tabular}
    \end{minipage}

    \caption{BEV predictions at the cross road in the east of \textit{Town02} when CVT~6ch.~KDE and CVT~6ch. fail to make a left turn.
Predictions at the same timestep.}
    \label{fig:cruzamento_4_town02}
\end{figure}

A qualitative analysis of the BEV predictions provides further insight into the driving failures observed in Table~\ref{tab:spawn_route_completion}. Figure~\ref{fig:curva_3_town01} illustrates a curve in Town01 where only \textit{CVT 6ch.} and \textit{CVT 6ch.\ KDE} produce a route channel that correctly traces the curvature ahead, allowing those agents to negotiate the turn, while \textit{CVT 3ch.} fails to represent the curve and the agent consequently cannot initiate the maneuver. The fact that both six-channel models succeed while \textit{CVT 3ch.} fails points to the larger and more diverse training dataset as the primary factor: steering and throttle perturbations expose both models to a wider range of trajectories through curves and off-nominal road positions. At this snapshot, the route channel (ch.\,1) IoU is 0.586 for \textit{CVT 6ch.\ KDE} and 0.669 for \textit{CVT 6ch.}, a ${\approx}$12\% difference that nonetheless leaves both predictions sufficient to guide the policy through the turn. The KDE weighting shows a stronger effect on the lane-boundary channel (ch.\,2), where its IoU of 0.203 represents a ${\approx}$283\% increase over the vanilla model's 0.053, illustrating that the KDE's benefit at this location is complementary rather than the decisive factor.

Figure~\ref{fig:cruzamento_1_town02} shows a left-turn intersection in Town02 where all agents fail. The better-performing models produce a route channel that tends to continue straight rather than tracing the intended left turn; \textit{CVT 6ch.\ KDE} yields the closest approximation among them, yet its prediction remains insufficient to signal the required direction change to the policy.

Another qualitative example is shown in Fig.~\ref{fig:cruzamento_4_town02}, focusing on a left-turn intersection in the eastern section of Town02, where at the time of the snapshot only \textit{CVT 6ch.\ KDE} and \textit{CVT 6ch.} were still active. Both agents fail to execute the left turn at this location, and the route channel predictions reveal a clear reason: the snapshot-level IoU for the route channel (channel1) is 0.156 for \textit{CVT 6ch.\ KDE} and only 0.092 for \textit{CVT 6ch.}, both values substantially below what would be needed to reliably convey the intended turning direction to the policy. The degraded route representation at this intersection means the agent receives an ambiguous or incorrect signal regarding the planned path, causing it to continue straight or commit an infraction rather than turn. This case reinforces the observation that low route channel accuracy at geometrically critical points, particularly intersections requiring directional changes, is a primary driver of episode terminations, and that improving BEV prediction quality at these locations remains the key challenge for enabling the policy to complete full routes under predicted BEV inputs. Given the critical role of the route channel in conveying directional intent to the policy, future work will focus specifically on improving its representation fidelity, as gains in this channel are expected to translate most directly into improved agent drivability.


\section{Conclusion}
\label{sec:conclusion}
The results presented in this work demonstrate that the KDE-weighted six-channel CVT model represents a meaningful step forward in BEV-based behavioral cloning for autonomous driving, being the only predicted-BEV agent capable of completing a full episode without infractions in the evaluated environment. The KDE weighting strategy proved effective in directing the model's learning capacity towards the most challenging and underrepresented driving scenarios, particularly curves and intersections, producing route and road channel predictions that are qualitatively more reliable. Nevertheless, the overall driving performance remains far below that of the ground-truth BEV baseline, indicating that the quality of the route channel prediction is still the primary bottleneck limiting closed-loop performance. Improving the fidelity of the route representation at intersections and sharp turns is therefore the most pressing direction for future work, as the current accuracy levels are insufficient to consistently guide the policy through directional maneuvers.

Once the route channel quality reaches a level that enables more reliable navigation of the road geometry, subsequent efforts should shift towards the obstacle-related channels, namely vehicles, pedestrians, and traffic lights, which were deliberately excluded from the scope of this evaluation. The near-zero IoU values observed for these channels across all models confirm that their prediction remains an open challenge, one that will become increasingly relevant as the evaluation is extended to environments with dynamic agents. In this broader context, the KDE weighting technique has demonstrated its value as a general mechanism for rebalancing the training signal towards infrequent but safety-critical scenarios, and exploring its application to the obstacle channels, potentially conditioned on the density of dynamic agent occurrences in the dataset rather than on ego-vehicle actions, represents a promising direction for future investigation.

\section*{Acknowledgment}
This work was partially funded by the National Council for Scientific and Technological Development – CNPq, Brazil (Grant No. 420148/2025-6), and by the Coordenação de Aperfeiçoamento de Pessoal de Nível Superior – Brasil (CAPES) – Finance Code 001.

%
%
%
\bibliographystyle{sbc}
\bibliography{biblio}

\end{document}